\documentclass[10pt,twocolumn,letterpaper]{article}

\usepackage{cvpr}              

\usepackage{booktabs} 
\usepackage{float}    
\usepackage{tabularx}
\usepackage{array}
\usepackage{dblfloatfix}
\usepackage{makecell}
\usepackage{mdframed}
\usepackage{tcolorbox}
\usepackage{epstopdf}
\tcbuselibrary{skins,breakable}

\newcommand{\TODO}[1]{\textbf{\color{red}[TODO: #1]}}

\newcolumntype{C}[1]{>{\centering\arraybackslash}p{#1}}
\renewcommand{\TODO}[1]{}

\usepackage{caption}
\definecolor{cvprblue}{rgb}{0.21,0.49,0.74}
\definecolor{applegreen}{rgb}{0.55, 0.71, 0.0}
\definecolor{ao}{rgb}{0.0, 0.5, 0.0}
\usepackage[pagebackref,breaklinks,colorlinks,allcolors=cvprblue]{hyperref}

\usepackage{kotex}

\def\paperID{} 
\def\confName{CVPR}
\def\confYear{2026}

\title{First Logit Boosting: Visual Grounding Method to Mitigate Object Hallucination in Large Vision-Language Models}

\author{Jiwoo Ha \quad Jongwoo Baek \quad Jinhyun So\thanks{Corresponding author} \\
DGIST EECS\\
{\tt\small djiwoo20@dgist.ac.kr, jaden9420@dgist.ac.kr, jinhyun@dgist.ac.kr}
}
\begin{document}
\maketitle
\begin{abstract}
Recent Large Vision-Language Models (LVLMs) have demonstrated remarkable performance across various multimodal tasks that require understanding both visual and linguistic inputs. However, object hallucination — the generation of nonexistent objects in answers — remains a persistent challenge. Although several approaches such as retraining and external grounding methods have been proposed to mitigate this issue, they still suffer from high data costs or structural complexity.
Training-free methods such as Contrastive Decoding (CD) are more cost-effective, avoiding additional training or external models, but still suffer from long-term decay, where visual grounding weakens and language priors dominate as the generation progresses. In this paper, we propose First Logit Boosting (FLB), a simple yet effective training-free technique designed to alleviate long-term decay in LVLMs. FLB stores the logit of the first generated token and adds it to subsequent token predictions, effectively mitigating long-term decay of visual information. We observe that FLB (1) sustains the visual information embedded in the first token throughout generation, and (2) suppresses hallucinated words through the stabilizing effect of the “The” token. Experimental results show that FLB significantly reduces object hallucination across various tasks, benchmarks, and backbone models. Notably, it causes negligible inference overhead, making it highly applicable to real-time multimodal systems. {Code is available at \url{https://github.com/jiwooha20/FLB}}
\end{abstract}

\section{Introduction}
\label{sec:intro}
Large Vision-Language Models (LVLMs) have recently achieved remarkable performance across a wide range of multimodal tasks that require joint understanding of images and text, such as image captioning and Visual Question Answering (VQA)~\cite{dai2023instructblip, liu2024improved, zhu2023minigpt, wang2024qwen2, chen2023shikra, ye2023mplug}.
Despite their impressive progress, LVLMs still suffer from a critical limitation known as object hallucination -- the generation of objects that are not present in the input image~\cite{liu2024survey, bai2024hallucination, li2023evaluating, lovenia2024negative}.
This phenomenon not only reduces model reliability, but also causes serious risks in safety-critical applications, including autonomous driving~\cite{chen2024driving, wu2023embodied} and medical imaging~\cite{hu2023advancing, wang2024interactive}.
Hence, addressing object hallucination remains essential for advancing LVLMs toward reliable real-world deployment.

\begin{figure}[t]
\vspace{-8pt}
    \centering
    \includegraphics[width=1\linewidth]{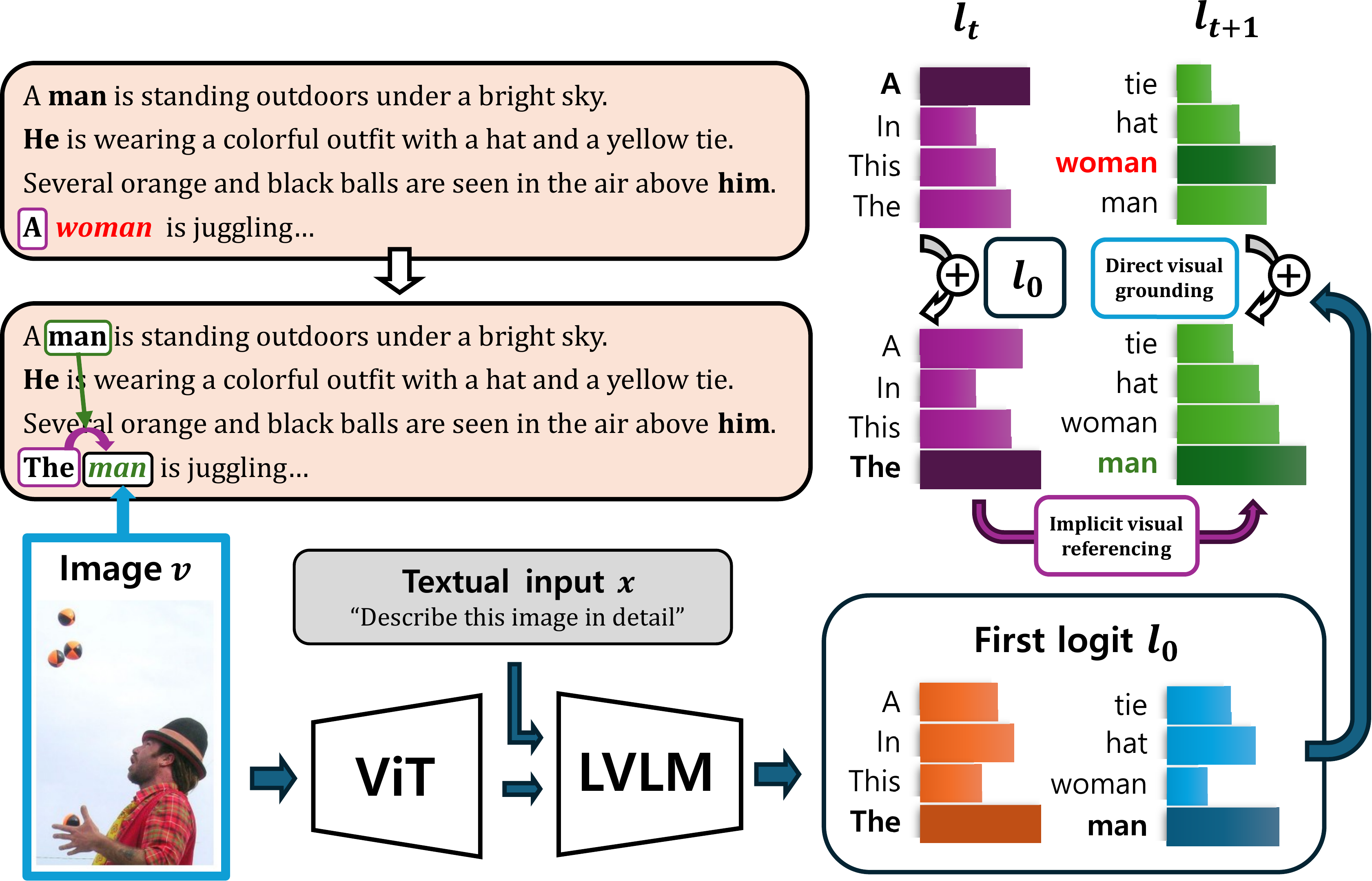}
    \caption{
    \textbf{Overview of First Logit Boosting (FLB).}
    FLB stores the logit of the first generated token and reuses it during decoding, which leverages two complementary effects. 
    (1) \textbf{Direct visual grounding}: the first token logit inherently carries stronger visual evidence (\textcolor{ao}{man}) than hallucinated token (\textcolor{red}{women}), serving as an anchor that preserves visual cues weakened by positional drift. 
    (2) \textbf{Implicit visual referencing}: by boosting the probability of starting a sentence with ``The'', FLB increases the likelihood of selecting nouns established before long-term decay occurs, thus maintaining referential coherence and mitigating hallucination.
    }
    \label{fig:overview}
\end{figure}

To mitigate this issue, existing studies can be broadly categorized into three lines of research.
(1) \textbf{Retraining-based methods} mitigate hallucination through additional training, such as Reinforcement Learning from Human Feedback (RLHF)~\cite{sun2024aligning, zhao2023beyond, gunjal2024detecting} on large-scale annotated datasets, or by architectural modifications such as positional embedding adjustments~\cite{xing2024mitigating}. 
(2) \textbf{External grounding methods} reduce hallucination by additional models that verify the existence of generated objects or supplement the missing ones~\cite{yin2024woodpecker,min2025mitigating, deng2024seeing}. 
(3) \textbf{Training-free methods} intervene directly in the logit or latent space at inference time, aiming to suppress hallucination without additional data or model components~\cite{Leng_2024_CVPR,liu2024paying,Huang_2024_CVPR,woo2025don,favero2024multi,wang2024mitigating, zhao2025cross, zhuang2025vasparse}. 
Nevertheless, all three lines exhibit inherent trade-offs: retraining methods are data-intensive and cost-intensive, external ground methods compromise efficiency, and existing training-free methods fail to sustain visual grounding over long generations, leading to persistent hallucinations.
Among the three methods, training-free approaches have recently gained attention for their efficiency and simplicity, as they operate entirely during the inference time without additional supervision or model retraining. 
Contrastive Decoding (CD), a representative training-free method, suppresses hallucination by contrasting the logits from the original and perturbed inputs, thereby reducing over-reliance on language priors and statistical biases~\cite{Leng_2024_CVPR, wang2024mitigating, zhao2025cross, favero2024multi, liu2024paying}.
However, these methods fail to fully address a critical issue, \emph{long-term decay} of visual grounding~\cite{min2025mitigating, favero2024multi}. 
As text generation progresses, the model's attention gradually drifts away from visual evidence and becomes dominated by language priors, causing hallucinations to frequently appear in later tokens. 
Moreover, CD methods require two forward passes per decoding step, one for the original input and the other for the perturbed version, which doubles inference time.

To overcome these limitations, we propose a simple yet effective training-free technique, named \textbf{First Logit Boosting (FLB)}, motivated by the key observation that the first token's logit inherently encodes the most visually grounded representation in LVLMs.
{Specifically, at the first decoding step, the relative logit gap between the ground truth and hallucination tokens is the largest.}
By storing the first token logit and reusing it in subsequent decoding steps, FLB continuously reinforces early visual grounding throughout the entire generation process.
As illustrated in \cref{fig:overview}, FLB leverages two complementary mechanisms that jointly stabilize decoding: (1) a \textbf{direct visual grounding effect} where the first token logit serves as a persistent visual anchor retaining visual evidence, and (2) an \textbf{implicit visual referencing effect} 
where the increased likelihood of sentence starting with ``The'' token encourages the model refer back to entities generated earlier before long-term decay becomes severe, thereby reducing the chance of hallucination. 
These dual effects effectively mitigate the long-term decay problem without introducing additional training and external models.
Moreover, unlike previous training-free methods such as CD, which require dual forward passes per decoding step, FLB achieves superior hallucination mitigation using only a single forward pass with negligible inference overhead. 
Our main contributions are as follows:
\begin{itemize}
    \item We propose FLB, a lightweight and training-free technique that effectively mitigates object hallucination caused by long-term decay without requiring additional training or external modules.
    \item We identify and analyze that FLB's improvement stems from two complementary effects, a direct visual grounding effect and an implicit visual referencing effect.
    \item Extensive experiments demonstrate that FLB achieves state-of-the-art hallucination mitigation performance across various benchmarks and backbone models, while preserving near-baseline inference efficiency. 
\end{itemize}
\section{Related work} \label{sec:related}

\subsection{Large Vision-Language Models}
LVLMs~\cite{dai2023instructblip, liu2024improved, zhu2023minigpt, wang2024qwen2} typically consist of three components: 
a vision encoder (e.g., CLIP~\cite{radford2021learning}), a Large Language Model (LLM) (e.g., LLaMA~\cite{touvron2023llama}, Vicuna~\cite{vicuna2023}) and cross-modal alignment module such as the Q-Former in InstructBLIP~\cite{dai2023instructblip} or the Multi-Layer Perceptron (MLP) projector in LLaVA-1.5~\cite{liu2024improved}. 
This architecture enables LVLMs to effectively integrate multimodal inputs and achieve strong performance on tasks like image captioning and VQA~\cite{hossain2019comprehensive}. 
However, they still suffer from object hallucination, where generated captions mention non-existent objects~\cite{liu2024survey, bai2024hallucination}.
\subsection{Mitigating hallucination}
Object hallucination in LVLMs arises from multiple causes, and consequently, various mitigation approaches have been explored.
Existing approaches can be roughly categorized into three groups:  
(1) training-based methods~\cite{sun2024aligning, zhao2023beyond}, 
(2) external grounding methods~\cite{yin2024woodpecker, min2025mitigating, deng2024seeing}, and 
(3) training-free methods~\cite{Leng_2024_CVPR, liu2024paying, Huang_2024_CVPR}.  
Specifically, training-free approaches typically operate on the logit or latent space during decoding to suppress language priors,
the tendency of LVLMs to rely more on textual context than on visual input, without additional models or retraining.  
Among these, CD methods mitigate hallucination by contrasting logits from the original input with those from a distorted input that amplifies language priors.
Different CD variants use distorted inputs such as a noised image~\cite{Leng_2024_CVPR}, a perturbed instruction~\cite{wang2024mitigating}, or an unconditioned input~\cite{favero2024multi}.

Despite their effectiveness, prior studies~\cite{min2025mitigating} have shown that the influence of language priors grows with sentence length, and existing CD methods fail to fully address this issue.  
Moreover, CD methods require additional decoding passes.
To overcome these limitations, we propose a training-free method 
which eliminates additional inference passes 
while alleviating the reinforcement of language priors that intensify with longer generations.
{Unlike previous logit steering methods that steer generation toward a target distribution, FLB instead leverages the earliest decoding distribution, which is the most visually grounded, with the ``The'' effect reflecting stabilization from reduced uncertainty.}

\section{Preliminary}
\label{sec:preliminary}
\subsection{LVLM decoding}
The LVLM parameterized by $\theta$ consists of a vision encoder and a language model.
The vision encoder processes the visual input $v$ and produces a set of visual features, 
which are fused with a text prompt $x$ to generate a textual response $y$.
The response is generated in an auto-regressive manner, 
where each token $y_t$ at step $t$ is sampled according to the following equation:
\begin{equation}
    y_t \sim p_{\theta}(y_t \mid v, x, y_{<t}) \propto \exp\big( \text{logit}_{\theta}(y_t \mid v, x, y_{<t}) \big),
    \label{eq:LVLM}
\end{equation}
where ${y}_{<t}$ denotes the sequence of previously generated tokens before time step $t$.
$\text{logit}_{\theta}(\cdot)$ represents the score that the model assigns to each candidate token, 
which is normalized into a probability distribution via the softmax function.
When the model assigns high probabilities to tokens that are not grounded in the visual input
-- often due to excessive reliance on language priors --
object hallucination occurs in the generated output.

\subsection{Contrastive decoding} \label{subsec:CD}
The CD method was proposed to mitigate object hallucinations in LVLMs.
CD methods generate two output logits: one from original input and another from a distorted input, which induces more stronger language prior. 
By contrasting these two logits, CD methods suppress over-reliance on language prior and thereby reduce hallucinations.
Among several variants, Visual Contrastive Decoding (VCD)~\cite{Leng_2024_CVPR} utilizes negative logit, which is from distorted image input $v'$.
The adjusted logits are computed by
\begin{align}
p_{\text{\tiny VCD}}(y \mid v, v', x)
= &\operatorname{softmax}\!\Big[(1+\alpha)\,\mathrm{logit}_{\theta}(y \mid v, x) \notag \\
&-\alpha\,\mathrm{logit}_{\theta}(y \mid v', x)\Big] \label{eq:PVCD}
\end{align}
where $\alpha$ is a parameter to determine the amplification of differences between the two distributions.

\begin{figure}[t]
  \centering
   \includegraphics[width=1\linewidth]{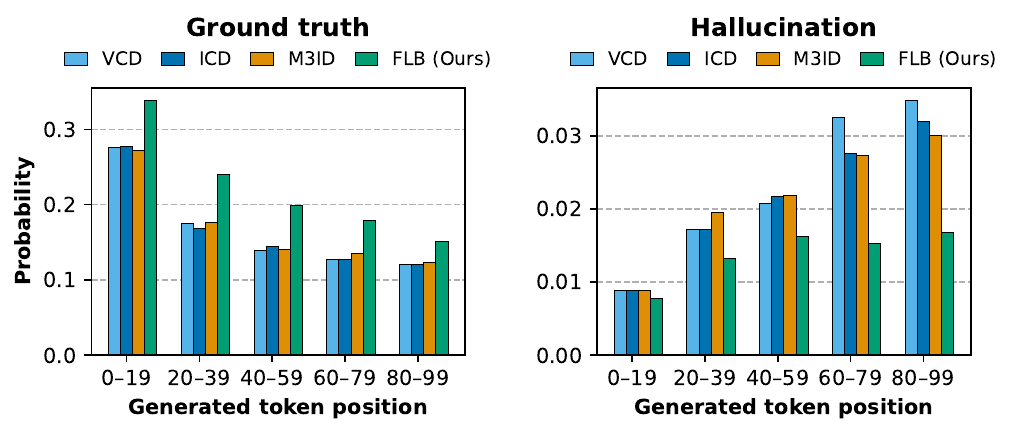}
   \caption{
    Comparison of the probability between all ground truth (left) and hallucination (right) words
    across token steps for each mitigation method. 
    As sentence length increases, hallucinated word logits become more dominant; while VCD, ICD, and M3ID fail to suppress this trend, FLB (ours) effectively mitigates hallucinated predictions.}
   \label{fig:step_cd}
     \vspace{-10pt}
\end{figure}
\subsection{The limitation of contrastive decoding} \label{subsec:limit_CD}

CD methods provide a model-agnostic and training-free way to mitigate hallucination. 
While these methods are more computationally efficient than retraining- or ground-based methods, they still suffer from two fundamental limitations: (1) the long-term decay of visual grounding and (2) inference inefficiency due to dual forward passes. 

\noindent {\bf Long-term decay.} 
Previous research~\cite{min2025mitigating, favero2024multi} and our own observations consistently show that as generation progresses, the influence of language priors strengthens while visual grounding gradually decays. 
This phenomenon originates from the Rotary Positional Embedding (RoPE)~\cite{su2024roformer} used in most LVLMs, where image tokens are placed at the beginning of the sequence. 
As the generation progresses, the relative positional distance between image and textual tokens increases, weakening cross-modal attention~\cite{xing2024mitigating}.
To empirically confirm this problem, we implement representative CD-based methods -- VCD~\cite{Leng_2024_CVPR}, Instruction Contrastive Decoding (ICD)~\cite{wang2024mitigating}, and Multi-Modal Mutual-Information Decoding (M3ID)~\cite{favero2024multi}. 
We introduce a conditional probability metric conditioned on noun probability to analyze how ground truth and hallucination related logits evolve across decoding steps, as shown in \cref{fig:step_cd}.
The results clearly show that, for all CD variants, the average logit of ground truth tokens consistently decreases while that of hallucinated tokens increases as the sequence length grows, demonstrating that none of these methods effectively addresses the long-term decay problem.

\noindent {\bf Inference inefficiency.}
Although CD-based methods avoid retraining, they require two forward passes per decoding step to obtain the original and contrastive logits. 
This design roughly doubles the inference time, and the overhead grows linearly with sequence length, which will be demonstrated in \cref{subsec:inf_spd}.

In summary, existing CD-based methods mitigate hallucination locally at each decoding step but fail to preserve visual grounding globally across long sequences. Moreover, their computational overhead at inference time undermines the practical advantages of being training-free. Addressing these two limitations motivates the design of our method introduced in the following section.
\renewcommand{\bottomfraction}{0.9}
\renewcommand{\textfraction}{0.1}

\section{Method}

\subsection{First Logit Boosting}

To mitigate the amplification of language priors in LVLMs during long sequences, we propose a simple yet effective technique named First Logit Boosting (FLB).
Our key intuition originates from the positional characteristics of LVLMs: the first token is generated immediately after visual tokens, before cross-modal attention begins to decay.
Hence, the first token's logit is minimally affected by the decay and therefore preserves the original visual information most faithfully.
Specifically, we store the logits of the first generated token given the prompt $x$ and visual input $v$: 
\begin{equation}
l_0 = \mathrm{logit}_{\theta}(y \mid x, v),
\label{eq:FL}
\end{equation}
and reuse it throughout subsequent decoding steps.
As the logit in~\cref{eq:FL} remains constant across decoding steps $t$,
the computation needs to be performed only once, incurring negligible additional cost.

Formally, FLB is formulated as follows:
\begin{equation}
\begin{split}
y_t \sim
 \operatorname{softmax}\!\Big[\mathrm{logit}_{\theta}(y \mid v, x, y_{<t}) +\,w_{t}\,l_0\Big],
\end{split}
\label{eq:FLB1}
\end{equation}
where $w_t$ is a temporal weighting function that gradually increases with decoding step $t$ to counteract long-term decay explained in~\cref{subsec:limit_CD}.
Specifically, $w_t$ is given by
\begin{equation} \label{eq:FLB2}
    w_{t} = \gamma (1 - e^{-\lambda t}),
\end{equation}
where $\gamma$ denotes the maximum scaling coefficient and $\lambda$ controls the rate of change.
Details on weight function and parameter optimization will be provided in~\cref{supp:hyp_opt}.

Since the first token logit does not correspond to the current decoding step, 
indiscriminate addition may increase the probability of selecting implausible tokens.
Therefore, we prevent this by applying an adaptive plausibility constraint, 
similar to that used in VCD~\cite{Leng_2024_CVPR}, to filter out tokens with low plausibility scores during prediction. 
The candidate set is defined as:
\begin{align}
\mathcal{V}_{\mathrm{head}}&(y_{<t})
= \{\, y_t \in \mathcal{V} : \notag \\
\!p_{\theta}(y_t &\mid v, x, y_{<t})
\ge \beta \max_{w} p_{\theta}(w \mid v, x, y_{<t}) \,\}, \\
p_{\text{\tiny FLB}}&(y_t \mid v, x)
= 0, \quad \text{if } y_t \notin \mathcal{V}_{\mathrm{head}}(y_{<t}), \notag
\label{eq:candidate}
\end{align}

where $\mathcal{V}$ denotes the output vocabulary of LVLM and $\beta \in [0,1]$ is a hyperparameter that determines the threshold for the candidate set $\mathcal{V}_{head}$, based on the maximum probability in the original logit distribution.
Finally, the full decoding rule is given by:
\begin{equation}
\begin{split}
y_t \sim
 \operatorname{softmax}\!\Big[\mathrm{logit}_{\theta}(y \mid x, v, y_{<t})  +\,w_{t}\,l_0\Big], \\
\text{subject to } y_t \in \mathcal{V}_{\mathrm{head}}(y_{<t}).
\end{split}
\label{eq:final}
\end{equation}

Through this mechanism, FLB obtains two complementary effects: (1) a direct visual grounding effect, where the stored first logit serves as a persistent anchor preserves visual consistency across decoding steps; and (2) an implicit visual referencing effect, named \emph{``The'' effect}, where the linguistic nature of the first token (e.g., ``The'') stabilizes subsequent noun predictions by encouraging references to previously grounded entities.
The following sections analyze these two effects in detail.

\begin{figure}[t]
  \centering
   \includegraphics[width=1\linewidth]{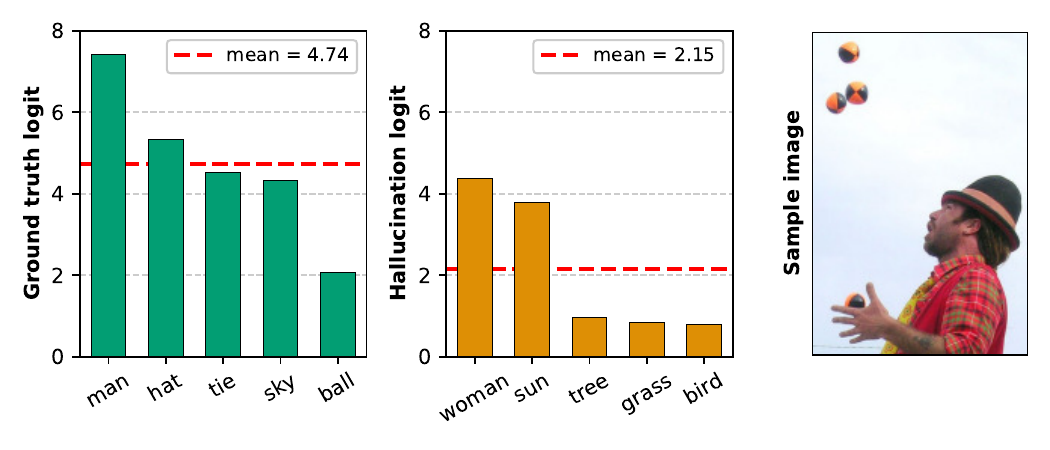}
   \vspace{-20pt}
   \caption{
   This figure shows the logits of the ground truth words (left) and hallucination words (middle) for the first token during caption generation for a case image (right). 
   The logit of ground truth words are generally higher than that of hallucination words. 
   }
   \label{fig:case}
\end{figure}


\subsection{Direct visual grounding effect} \label{subsec:direct_effect}

The first effect of FLB directly addresses the long-term decay of visual grounding.
As discussed in~\cref{subsec:limit_CD}, LVLMs suffer from the long-term decay due to RoPE, which weakens cross-modal attention as the distance between image and text tokens increases.
This section investigates how FLB directly addresses this issue by leveraging the logit of the first token as a persistent visual anchor.
To quantitatively examine the effect, we analyze the logit difference between the ground truth and hallucination tokens contained in the first logit in~\cref{eq:FL}. As shown in~\cref{fig:case}, the ground truth words exhibit consistently higher logits than the hallucinated ones, demonstrating that the first token retains the strongest visual correlation.
As FLB reuses this logit across all subsequent decoding steps, this margin is repeatedly applied, continuously reinforcing visual signals.

\begin{figure}[t]
  \centering
   \includegraphics[width=0.9\linewidth]{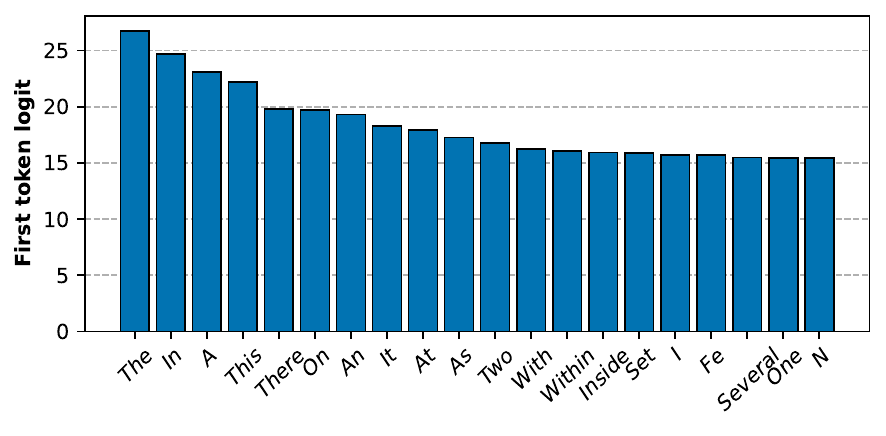}
   \vspace{-10pt}
   \caption{Top 20 tokens by logit value for the first token prediction. The list includes common sentence-starting words such as ``The'', ``In'', and ``A''.}
   \label{fig:first}
   \vspace{-12pt}
\end{figure}

\subsection{``The'' effect} \label{subsec:the_effect}

While FLB is originally designed to mitigate visual decay through the direct visual grounding effect, we unexpectedly observe a complementary effect arising from the linguistic properties of the first token.
As illustrated in~\cref{fig:first}, the first decoding step generates articles such as ``The'', ``In'', and ``A'', which subtly influence the model's subsequent predictions. 
Surprisingly, sentences beginning with ``The'' exhibit markedly lower hallucination rates, and nouns following ``The'' are more frequently grounded in visual input than those following ``A'' or the other starting tokens.
This emergent effect, named \emph{``The'' effect}, reveals that linguistic initialization can act as an implicit form of visual grounding. 
Even without any modification to the model's language prior, FLB amplifies this bias toward contextually and visually grounded nouns by reusing the first token logit, leading to more coherent and visually consistent generations. 
We empirically validate this effect in~\cref{subsec:ablation} where we provide a detailed analysis of (i) sentences starting with ``The'', (ii) nouns immediately following ``The'', and (iii) the reduction in prediction entropy after ``The'', jointly offering a comprehensive account of the \emph{``The'' effect}.
\vspace{-5pt}
\begin{table*}[t]
  \caption{Performance comparison of LLaVA1.5 and InstructBLIP on AMBER generative tasks. The \textbf{highest} scores are marked in \textbf{bold}.}
  \label{tab:amber_result}
  \centering
  \small
  \resizebox{\textwidth}{!}{%
    \begin{tabularx}{\textwidth}{
    >{\centering\arraybackslash}X|
    >{\centering\arraybackslash}X>{\centering\arraybackslash}X
    >{\centering\arraybackslash}X>{\centering\arraybackslash}X|
    >{\centering\arraybackslash}X>{\centering\arraybackslash}X
    >{\centering\arraybackslash}X>{\centering\arraybackslash}X
  }  
    \toprule\toprule
    \multicolumn{1}{c}{\textbf{AMBER}} &
    \multicolumn{4}{c}{\textbf{LLaVA1.5}} &
    \multicolumn{4}{c}{\textbf{InstructBLIP}} \\
    \cmidrule(lr){2-5} \cmidrule(lr){6-9}
    \textbf{Method} & \textbf{CHAIR$\downarrow$} & \textbf{Cover$\uparrow$} & \textbf{Hal$\downarrow$} & \textbf{Cog$\downarrow$}
                   & \textbf{CHAIR$\downarrow$} & \textbf{Cover$\uparrow$} & \textbf{Hal$\downarrow$} & \textbf{Cog$\downarrow$} \\
    \midrule
    Baseline & $11.5\,(\pm\,0.29)$ & $50.1\,(\pm\,0.51)$ & $48.9\,(\pm\,0.78)$ & $4.6\,(\pm\,0.16)$
            & $11.6\,(\pm\,0.22)$ & $53.4\,(\pm\,0.12)$ & $51.7\,(\pm\,1.73)$ & $5.3\,(\pm\,0.29)$ \\
    VCD     & $ 9.9\,(\pm\,0.12)$ & $51.2\,(\pm\,0.24)$ & $43.4\,(\pm\,0.37)$ & $4.6\,(\pm\,0.08)$
            & $10.2\,(\pm\,0.05)$ & $53.5\,(\pm\,0.33)$ & $46.9\,(\pm\,1.09)$ & $4.8\,(\pm\,0.21)$ \\
    ICD     & $ 9.1\,(\pm\,0.29)$ & $51.2\,(\pm\,0.05)$ & $40.6\,(\pm\,0.98)$ & $4.3\,(\pm\,0.12)$
            & $12.1\,(\pm\,0.61)$ & $52.6\,(\pm\,0.17)$ & $51.4\,(\pm\,0.86)$ & $5.2\,(\pm\,0.21)$ \\
    M3ID    & $ 9.8\,(\pm\,0.22)$ & $\textbf{55.6}\,(\pm\,0.25)$ & $48.4\,(\pm\,1.11)$ & $3.6\,(\pm\,0.09)$
            & $ 11.5\,(\pm\,0.61)$ & $ 52.5\,(\pm\,0.17)$ & $ 51.4\,(\pm\,1.28)$ & $ \textbf{4.6}\,(\pm\,0.24)$ \\
    FLB (Ours)    & $\textbf{6.1}\,(\pm\,0.37)$  & $50.4\,(\pm\,0.22)$ & $\textbf{31.6}\,(\pm\,0.99)$  & $\textbf{2.7}\,(\pm\,0.24)$
            & $\textbf{ 9.0}\,(\pm\,0.19)$ & $\textbf{53.6}\,(\pm\,0.26)$ & $\textbf{43.8}\,(\pm\,1.13)$  & $4.7\,(\pm\,0.14)$ \\
    \bottomrule\bottomrule
  \end{tabularx}%
  }
\end{table*}
\begin{table*}[t]
  \caption{Performance comparison of LLaVA1.5 and InstructBLIP on CHAIR generative tasks.}
  \label{tab:chair_result}
  \centering
  \small
  \begin{tabularx}{\textwidth}{
    >{\centering\arraybackslash}X|
    >{\centering\arraybackslash}X>{\centering\arraybackslash}X
    >{\centering\arraybackslash}X|
    >{\centering\arraybackslash}X>{\centering\arraybackslash}X
    >{\centering\arraybackslash}X
  } 
    \toprule\toprule
    \multicolumn{1}{c}{\textbf{CHAIR}} &
    \multicolumn{3}{c}{\textbf{LLaVA1.5}} &
    \multicolumn{3}{c}{\textbf{InstructBLIP}} \\
    \cmidrule(lr){2-4} \cmidrule(lr){5-7}
    \textbf{Method} & \textbf{CHAIR\(_s\) \(\downarrow\)} & \textbf{CHAIR\(_i\) \(\downarrow\)} & \textbf{Recall \(\uparrow\)} 
                    & \textbf{CHAIR\(_s\) \(\downarrow\)} & \textbf{CHAIR\(_i\) \(\downarrow\)} & \textbf{Recall \(\uparrow\)}  \\
    \midrule
    Baseline & $57.5\,(\pm\,2.23)$ & $17.3\,(\pm\,0.74)$ & $73.3\,(\pm\,0.90)$ 
            & $59.0\,(\pm\,1.84)$ & $18.5\,(\pm\,0.37)$ & $69.4\,(\pm\,0.90)$  \\
    VCD & $57.0\,(\pm\,1.02)$ & $16.3\,(\pm\,0.85)$ & $\textbf{76.7}\,(\pm\,0.54)$ & $58.7\,(\pm\,0.19)$
            & $18.5\,(\pm\,0.47)$ & $69.3\,(\pm\,0.57)$  \\
    ICD & $53.0\,(\pm\,1.93)$ & $14.6\,(\pm\,0.85)$ & $\textbf{76.7}\,(\pm\,0.46)$ & $65.7\,(\pm\,0.77)$
            & $20.1\,(\pm\,0.34)$ & $\textbf{74.9}\,(\pm\,1.61)$  \\
    M3ID & $54.5\,(\pm\,0.68)$ & $15.9\,(\pm\,0.36)$ & $73.5\,(\pm\,0.39)$ & $69.8\,(\pm\,2.30)$
            & $21.4\,(\pm\,1.10)$ & $70.8\,(\pm\,0.38)$ \\
    FLB (Ours) & $\textbf{43.5}\,(\pm\,1.24)$ & $\textbf{12.0}\,(\pm\,0.50)$ & $73.6\,(\pm\,0.48)$ & $\textbf{52.5}\,(\pm\,0.47)$
            & $\textbf{15.8}\,(\pm\,0.86)$ &  $71.3\,(\pm\,0.49)$ \\

    \bottomrule\bottomrule
  \end{tabularx}
  \vspace{-10pt}
\end{table*}
\section{Experiment}
\vspace{-2pt}
In this section, we demonstrate the effectiveness of the proposed FLB in mitigating hallucination.
Since FLB is designed to suppress the influence of language priors that become stronger as sentences grow longer, 
its effectiveness in mitigating hallucination increases accordingly. 
Therefore, we focused on \emph{generative} tasks, where such effects are more evident, rather than discriminative ones.
\subsection{Datasets and evaluation metrics}
\vspace{-2pt}
We conducted main experiments on two standard benchmarks that require long and descriptive caption generation:
the CHAIR~\cite{rohrbach2018object} and AMBER~\cite{wang2023amber} datasets. 
For CHAIR, we used 500 images randomly sampled from the MSCOCO 2014 validation set~\cite{lin2014microsoft}. 
For the AMBER benchmark, we used all 1,004 images from the AMBER dataset.
Captions were generated using the prompt on both benchmarks:  
\textit{``Please describe this image in detail.''}  
The generated captions were then evaluated using following metrics~\cite{rohrbach2018object, wang2023amber}:
(1) \textbf{CHAIR} (CHAIR\textsubscript{i}), the proportion of hallucinated objects among all generated objects;  
(2) \textbf{Cover}, the proportion of ground truth objects among all generated objects;  
(3) \textbf{Hal} (CHAIR\textsubscript{s}), the proportion of captions containing at least one hallucinated object; and  
(4) \textbf{Cog}, the proportion of hallucinated objects that are perceptually plausible to humans.

To further evaluate the generalization capability of FLB beyond captioning tasks, we additionally conducted experiments on diverse tasks and interaction styles.
We used two benchmarks: MMHalBench~\cite{sun2024aligning}, which includes diverse question types (e.g., object attributes, relations, and contextual reasoning) beyond image captioning, and ConvBench~\cite{liu2024convbench}, a three-turn conversation-style benchmark.
Both benchmarks are evaluated using GPT-4V as the judge.
\vspace{-12pt}
\subsection{MLLM backbones}
\vspace{-2pt}
To verify the generality of FLB, we evaluated the two benchmarks on LVLMs employing different types of visual–language alignment. 
Specifically, we used two representative models: 
\textbf{LLaVA-1.5 (7B)}~\cite{liu2024improved}, which connects vision and language modalities through an MLP-based projection layer, 
and \textbf{InstructBLIP (7B)}~\cite{dai2023instructblip}, which adopts a Q-former~\cite{li2023blip} architecture for cross-modal alignment.
\subsection{Baselines}
To compare FLB with existing training-free hallucination mitigation methods, 
we included \textbf{Baseline decoding} (vanilla decoding without any hallucination mitigation method) and three representative contrastive decoding techniques as baselines: 
\textbf{VCD}~\cite{Leng_2024_CVPR}, \textbf{ICD}~\cite{wang2024mitigating}, and \textbf{M3ID}~\cite{favero2024multi}.  
VCD alleviates hallucination by contrasting logits obtained from an original image and a distorted image, 
while ICD performs contrastive decoding using an original instruction and a perturbed instruction.
M3ID reduce hallucinations by minimizing visual information loss using image conditioned input and unconditioned input.
All experiments were conducted with random sampling-based decoding. 
Each result was averaged over three runs to ensure consistency.  
For each baseline, we adopted hyperparameter configurations that achieved the best performance as reported in the original papers~\cite{Leng_2024_CVPR, wang2024mitigating, favero2024multi}.
The hyperparameters of FLB were set to $\gamma = 0.3$, $\lambda = 0.05$, and $\beta = 0.1$,
as detailed in~\cref{supp:hyp_opt}.


\begin{table}[t]
  \caption{Average count of words and tokens in generated sequences tested on LLaVA1.5.}
  \label{tab:avg_length}
  \centering
  \small
  \begin{tabularx}{0.95\linewidth}{
    >{\centering\arraybackslash}X|
    >{\centering\arraybackslash}X
    >{\centering\arraybackslash}X
  }
    \toprule\toprule
    & \mbox{Average words} & \mbox{Average tokens} \\
    \midrule
    Baseline & 79.58 & 104.67 \\
    FLB & 78.62 & 101.40 \\
    \bottomrule\bottomrule
  \end{tabularx}
\end{table}
\begin{table}[t]
  \caption{GPT-4V-aided Evaluation Results.}
  \label{tab:GPT-4V}
  \centering
  \small
  \begin{tabularx}{0.95\linewidth}{
    >{\centering\arraybackslash}X|
    >{\centering\arraybackslash}X
    >{\centering\arraybackslash}X
  }
    \toprule \toprule
    & Accuracy & Detailedness \\
    \midrule
    Baseline & 5.01 & 5.47 \\
    FLB & \textbf{7.28} & \textbf{6.51} \\
    \bottomrule\bottomrule
  \end{tabularx}
  \vspace{-10pt}
\end{table}

\section{Result and analysis}
{
In this section, we assess the effectiveness of FLB.
In \cref{subsec:main_result,subsec:QoS,subsec:inf_spd,subsec:ablation}, we present the main comparison results, as well as analyses of generation quality, inference speed, and ablation studies on the AMBER and CHAIR datasets.
In \cref{subsec:beyond}, we further evaluate the robustness of FLB across diverse tasks and benchmarks.
}
\subsection{Main result} \label{subsec:main_result}
\cref{tab:amber_result} presents the average performance of FLB on the AMBER benchmark.
Across both LLaVA-1.5 and InstructBLIP, FLB outperforms all compared methods, surpassing the prior hallucination mitigation techniques, including VCD, ICD, and M3ID {on hallucination-related metrics}.
\cref{tab:chair_result} presents the results on the MSCOCO-based CHAIR generative benchmark, showing consistent gains across different model architectures and datasets.
These results collectively demonstrate the robustness and general applicability of FLB as a model-agnostic framework to mitigate generative hallucination in LVLMs.
{Importantly, these improvements are achieved without degrading performance on ground truth related metrics, 
effectively overcoming the typical trade-off between hallucination reduction and answer correctness.}

\subsection{Qualitative Evaluation} \label{subsec:QoS}

Since FLB directly adds logits from different token positions, 
this raises concerns that such cross-position manipulation could distort the natural language generation process and degrade sentence fluency 
(e.g., inserting capitalized tokens, such as ``The'' in the middle of a sentence).
To assess this risk, we conducted experiments to evaluate the impact of FLB on sentence quality.
First, we compared the average number of words and tokens in sentences generated on the AMBER benchmark in \cref{tab:avg_length}. We confirmed that there was no significant difference in sentence length or the number of tokens before and after applying FLB. 
Furthermore, across the entire AMBER dataset (approximately 1,000 sentences), no cases were observed in which unexpected capitalized tokens appeared mid-sentence.
This {complete absence of failure case} indicates that the adaptive plausibility constraint in FLB effectively suppresses excessive logit perturbations, thereby maintaining grammatical consistency.
Second, we adopted the GPT-4-based evaluation framework proposed by~\cite{yin2024woodpecker} to compare sentence-level quality.
As shown in \cref{tab:GPT-4V}, sentences generated with FLB achieved higher quality scores, confirming that FLB does not compromise grammaticality or expressive fluency.

\subsection{Inference speed} \label{subsec:inf_spd}
\begin{figure}
\vspace{-6pt}
  \centering
   \includegraphics[width=0.9\linewidth]{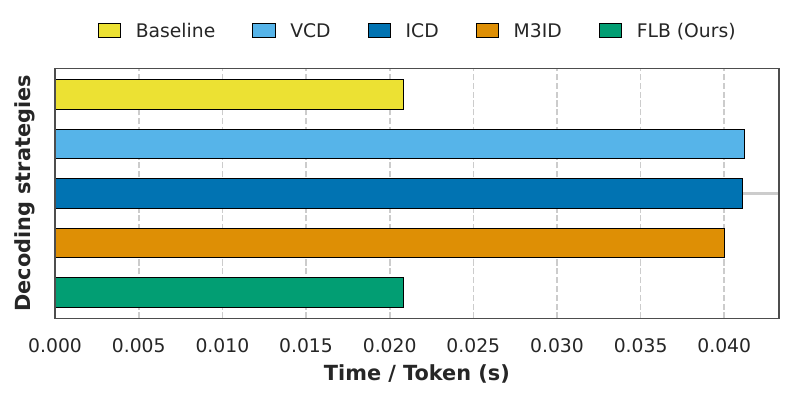}
    \vspace{-4pt}
    \caption{\textbf{Inference speed comparison across decoding strategies.} This speed is measured by the time it takes to generate each token. VCD/ICD/M3ID run about twice as slow as the baseline, while FLB (ours) maintains similar speed.}
   \label{fig:speed}
   \vspace{-10pt}
\end{figure}

We analyzed the impact of various hallucination mitigation methods and FLB on the inference speed of LVLMs.
To evaluate this, we measured the average generation speed on the AMBER generative task.
As shown in \cref{fig:speed}, CD methods exhibit nearly twice slower inference speed compared to the baseline model without any mitigation.
In contrast, FLB, which only reuses the saved logit of the first token,
shows no degradation in inference speed.
These results demonstrate that FLB achieves hallucination mitigation with negligible computational overhead, maintaining near-baseline efficiency and outperforming existing decoding-based methods that suffer from significant inference latency.

\subsection{Ablation study} \label{subsec:ablation}
To investigate the contribution of the two core mechanisms underlying FLB, the direct visual grounding effect and ``The'' effect', we conducted an ablation analysis designed to isolate each effect individually. 
\cref{tab:ablation_study} summarizes the results in four configurations:
(1) \textbf{Baseline}: vanilla LVLM without hallucination mitigation;
(2) \textbf{Direct visual grounding only}: we apply FLB while masking all logits in $l_0$ except those corresponding to \emph{noun tokens}, thereby activating only the direct visual grounding effect;
(3) \textbf{``The'' effect only}: we mask all logits in $l_0$ except the ``The'' token, which selectively increases the probability of generating sentences that begin with ``The'', isolating the implicit visual referencing effect;
(4) \textbf{FLB (full)}: the proposed full configuration, in which both effects operate jointly throughout decoding.
As shown in \cref{tab:ablation_study}, both isolated configurations outperform the baseline, confirming that each effect contributes a distinct benefit. 
The full FLB configuration achieves the best overall results, demonstrating that the two effects are complementary rather than redundant.
Interestingly, while the direct visual grounding effect achieves moderate gain by reinforcing visually aligned object predictions, ``The'' effect yields a larger gain, which motivates us to further investigate impact of the ``The'' token in the following subsections. 
\vspace{-13pt}
\begin{table}[t]
    \small
    \caption{Ablation results isolating the two core effects of FLB.}
    \label{tab:ablation_study}
    \centering
    \begin{tabularx}{1\linewidth}{
        >{\centering\arraybackslash}m{0.425\linewidth} |
        >{\centering\arraybackslash}m{0.11\linewidth}
        >{\centering\arraybackslash}m{0.09\linewidth}
        >{\centering\arraybackslash}m{0.06\linewidth}
        >{\centering\arraybackslash}m{0.06\linewidth}
    }
        \toprule\toprule
        & CHAIR\(\downarrow\) & Cover\(\uparrow\) & Hal\(\downarrow\) & Cog\(\downarrow\) \\
        \midrule
        Baseline & 11.9 & 49.6 & 48.8 & 4.4 \\
        \mbox{Direct visual grounding only}  & 9.2 & 50.3 & 41.1 &4.7 \\ 
        ``The'' effect only & 6.5 & 50.6 & 29.9 & 2.4 \\
        FLB (full) & 5.7 & 50.3 & 30.7 & 2.4 \\
        \bottomrule
        \bottomrule
    \end{tabularx}
\end{table}
\begin{table}[t]
  \caption{\textbf{Frequency statistics of token occurrences after articles.} The table reports the proportion (and raw counts) of ground truth and hallucination words appearing after the articles The/the and A/a.}
  \label{tab:article_stats}
  \centering
  \small
  \begin{tabularx}{0.95\linewidth}{
    >{\centering\arraybackslash}X|
    >{\centering\arraybackslash}X
    >{\centering\arraybackslash}X
  }
    \toprule\toprule
    & After The/the & After A/a \\
    \midrule
    Ground truth & 0.317 (2,424) & 0.359 (853) \\
    Hallucination & 0.020 (150) & 0.105 (248) \\
    \bottomrule\bottomrule
  \end{tabularx}
\end{table}
\begin{table}
  \caption{\textbf{Average prediction probabilities of token occurrences after articles.} The table presents the average model prediction probabilities for ground truth and hallucination tokens appearing after The/the and A/a.}
  \label{tab:word prob}
  \centering
  \small
    \begin{tabularx}{0.95\linewidth}{
    >{\centering\arraybackslash}X|
    >{\centering\arraybackslash}X
    >{\centering\arraybackslash}X
    }
    \toprule\toprule
    & After The/the & After A/a \\
    \midrule
    Ground truth & 0.279 & 0.225 \\
    Hallucination & 0.012 & 0.029 \\
    \bottomrule\bottomrule
  \end{tabularx}
  \vspace{-15pt}
\end{table}

\subsubsection{Sentence level analysis}
\begin{figure}[t]
\vspace{-10pt}
  \centering
   \includegraphics[width=1\linewidth]{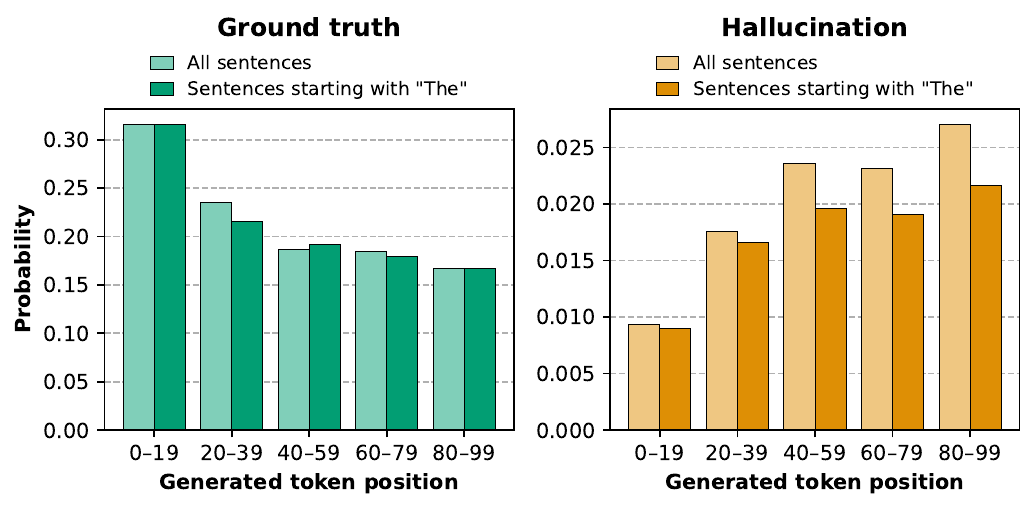}
   \caption{
    \textbf{Sentence-level impact of ``The'' on long-term hallucination.}
    We compare probability of ground truth (left) and  hallucination (right) words across generated token positions for two groups: 
    sentences beginning with ``The'' and all other sentences. 
    Sentences beginning with ``The'' show comparable ground truth probability to the overall distribution but exhibit substantially lower hallucination growth, particularly at later token positions where long-term decay intensifies. 
    The widening gap between two groups demonstrate that ``The'' token provides a stabilizing initialization that suppresses cumulative hallucination. 
   }
   \label{fig:step_the2}
   \vspace{-10pt}
\end{figure}

A key observation from \cref{subsec:the_effect} is that the first token logit frequently corresponds to sentence-initial articles with ``The'' having the highest logit (see \cref{fig:first}).
We hypothesize that ``The'' token reduces long-term hallucination by encouraging the model to refer {back to} entities that have already been visually attended to, rather than sampling novel or ambiguous objects influenced by language priors. 
To test this, we divide the captions generated by LLaVA-1.5 into two groups, sentences beginning with ``The'' and all others, 
and measure conditional probability of ground truth and hallucination nouns in the same way as in \cref{fig:step_cd}.
As shown in \cref{fig:step_the2}, hallucination probability increases sharply in ordinary sentences, reflecting long-term decay driven by language priors, but grows marginally in sentences beginning with ``The''. 
More importantly, the divergence between the two groups becomes larger at later token positions.
These results suggest that even without applying FLB, the presence of ``The'' at the beginning of a sentence inherently suppresses long-term decay and reduces hallucination,
possibly by stabilizing the model’s initial linguistic state and mitigating early-stage degradation of visual grounding.
\subsubsection{Word level analysis}
To verify whether this stabilizing effect of the ``The'' token extends to the nouns that follow it,
we conducted an experiment to examine whether nouns appearing after ``The'' are indeed less likely to be hallucinated.
Specifically, we compared sentences beginning with the tokens ``The'' and ``A,''
which both serve as articles preceding nouns, and analyzed the nouns following each token.
\cref{tab:article_stats} reports the frequency of ground truth and hallucination nouns following each token.
The results indicate that nouns following the ``The'' token exhibit a significantly lower hallucination rate than those following the ``A'' token.
Moreover, \cref{tab:word prob} compares the average prediction probabilities of these nouns,
revealing that nouns predicted after ``The'' have a considerably lower average probability of being hallucinated compared to those following ``A.''
This finding suggests that the ``The'' token helps stabilize noun prediction and suppress the likelihood of hallucination during subsequent word generation.

\subsubsection{Entropy analysis}
The above analyses demonstrate the influence of ``The'' token on the occurrence of object hallucination.
In this section, we further discuss the critical underlying factor that explains this effect: \emph{entropy reduction.}
Prior research found various indicators of hallucination in LLMs, 
and one of the most representative metric is entropy of the logits for next token prediction.
In other words, higher uncertainty of the next token prediction correlates with lower confidence of model and increased hallucination~\cite{farquhar2024detecting, kalai2025language}.


\begin{table}
  \caption{\textbf{Logit entropy statistics under different conditions.} The table compares average logit entropy across total tokens, AMBER ground truth versus hallucination nouns, and nouns following different articles. Numbers in parentheses denote sample counts.}
  \label{tab:entropy_stats}
  \centering
  \small
  \begin{tabularx}{1\linewidth}{
    >{\centering\arraybackslash}m{0.23\linewidth} |
    >{\centering\arraybackslash}m{0.35\linewidth}
    >{\arraybackslash}m{0.3\linewidth}
    }
    \toprule\toprule
    & & Entropy \\
    \midrule
    \textbf{Total} & \mbox{All tokens (total steps)} & $1.815\,(105,803)$ \\
                   & All nouns & $2.305\,(16,397)$  \\
    \midrule
    \textbf{AMBER} & Ground truth & 1.948 (2,949) \\
                   & Hallucination & 3.265 (589) \\
    \midrule
    \mbox{\textbf{After articles}} & After The/the & 2.001 (4,858) \\
                   & After other than The/the & 2.433 (11,539) \\
                   & After A/a & 3.190 (2,286) \\
    \bottomrule\bottomrule
  \end{tabularx}
\end{table}
\begin{table}
  \caption{Comparison of total noun entropy before and after applying FLB. Numbers in parentheses denote sample counts.}
  \label{tab:total_entropy}
  \centering
  \small
    \begin{tabularx}{0.95\linewidth}{
    >{\centering\arraybackslash}m{0.23\linewidth} |
    >{\centering\arraybackslash}m{0.17\linewidth}
    >{\centering\arraybackslash}m{0.17\linewidth}
    >{\centering\arraybackslash}m{0.17\linewidth}
  }
    \toprule\toprule
    & \makecell{Baseline} & \makecell{``The'' \\effect only} & \makecell{FLB (full)} \\
    \midrule
    \makecell{Entropy \\ (sample count)} & \makecell{2.305 \\ (16,397)} & \makecell{2.210 \\(15,215)} & \makecell{2.181 \\(15,648)} \\
    \bottomrule\bottomrule
  \end{tabularx}

\end{table}
\begin{table}
  \caption{{Performances of LLaVA1.5 on MMHalbench.}}
  \label{tab:mmHalbench}
  \centering
  \small
  \begin{tabularx}{0.95\linewidth}{
    >{\centering\arraybackslash}m{0.23\linewidth} |
    >{\centering\arraybackslash}m{0.17\linewidth}
    >{\centering\arraybackslash}m{0.17\linewidth}
    >{\centering\arraybackslash}m{0.2\linewidth}
  }
    \toprule\toprule
    & \makecell{Baseline} & \makecell{VCD} & \makecell{FLB (Ours)} \\
    \midrule
    \makecell{Average Score \\ (variance)}
      & \makecell{1.944 \\ ($\pm 0.12$)}
      & \makecell{2.098 \\ ($\pm 0.05$)}
      & \makecell{\textbf{2.230} \\ ($\pm 0.18$)} \\
    \bottomrule\bottomrule
  \end{tabularx}
\end{table}
\begin{table}
  \caption{{Performances of LLaVA1.5 on Convbench.}}
  \label{tab:convbench}
  \centering
  \small
  \begin{tabularx}{0.95\linewidth}{
    >{\centering\arraybackslash}m{0.2\linewidth} |
    >{\centering\arraybackslash}m{0.17\linewidth}
    >{\centering\arraybackslash}m{0.17\linewidth}
    >{\centering\arraybackslash}m{0.2\linewidth}
  }
    \toprule\toprule
    \makecell{Win rate}& \makecell{Baseline} & \makecell{VCD} & \makecell{FLB (Ours)} \\
    \midrule
    \makecell{1st turn}
      & \makecell{0.132}
      & \makecell{0.154}
      & \makecell{\textbf{0.159}} \\
    \makecell{2nd turn}
      & \makecell{0.173} 
      & \makecell{0.173}
      & \makecell{\textbf{0.178}} \\
    \makecell{3rd turn}
      & \makecell{0.103} 
      & \makecell{\textbf{0.111}}
      & \makecell{0.108} \\

    \bottomrule\bottomrule
  \end{tabularx}
  \vspace{-15pt}        
\end{table}

Motivated by this perspective, we measure the entropy of logits specifically when the model predicts nouns.
Concretely, we evaluate:
(1) the entropy of ground truth and hallucination nouns, 
and (2) the entropy of nouns that appear after specific articles, such as The or A.
As shown in \cref{tab:entropy_stats},
the result of (1) show that the model exhibits higher logit entropy when generating hallucinated nouns as in the case of prior research.
Interestingly, the analysis in (2) reveals that nouns following ``The'' token have significantly lower entropy compared to nouns following other articles.
This result indicates that ``The'' token stabilizes the initial prediction distribution and reduces linguistic-level uncertainty.
Additionally, \cref{tab:total_entropy} presents that both the ``The'' effect only setting and the full FLB substantially reduce noun-level entropy, indicating more stable and confident predictions.
In summary, the ``The'' token effectively reduces model uncertainty and improves sentence-level stability compared to other starting tokens, 
serving as a key factor in mitigating object hallucination in LVLMs.
\begin{figure}[t]
    \vspace{-10pt}
    \centering
    \includegraphics[width=1\linewidth]{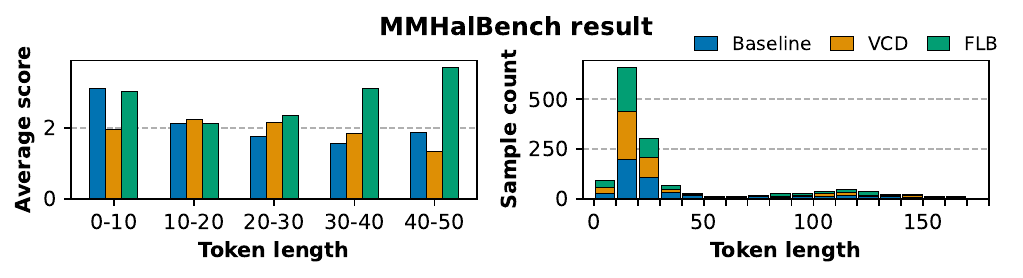}
    \vspace{-18pt}
    \caption{
    {\textbf{
    MMHalBench results across token positions for each method (left) and the distribution of generated token lengths (right).} Based on the token length distribution, we focus on sentences with 0–50 tokens and compare the average MMHalBench score. Unlike other methods, FLB maintains robust performance as sentence length increases.
    }}
    
    \label{fig:mmhal}
    \vspace{-10pt}
\end{figure}

\subsection{Beyond caption generation task} \label{subsec:beyond}
{
In this section, we evaluate FLB on tasks beyond standard caption generation, covering diverse tasks and interaction styles to assess its generalization capability. 
As shown in \cref{tab:mmHalbench}, FLB consistently outperforms both the baseline and VCD across five runs on MMHalbench.
Specifically, FLB achieves an average score improvement of 0.286 over the baseline, whereas VCD achieves an improvement of 0.154.
Moreover, analysis by token length shows that FLB’s advantage grows for longer outputs, 
while both baseline and VCD degrade, supporting our hypothesis that FLB mitigates long-term decay of visual grounding as shown in~\cref{fig:mmhal}.
Similarly, in \cref{tab:convbench}, FLB improves over the baseline by 0.027, 0.005, and 0.005 across the three turns, respectively on Convbench, respectively.
Overall, these results show that FLB preserves linguistic naturalness while suppressing hallucination, and generalizes as a decoding-time mechanism across diverse tasks.
}
\section{Conclusion}
In this paper, we introduced FLB, a lightweight and training-free decoding method that mitigates long-term object hallucination in LVLMs by reusing the first-token logit throughout generation. 
{Unlike prior approaches that introduce additional models or require double inference, 
FLB achieves hallucination mitigation through reusing only the first token logit.}
FLB leverages two complementary mechanisms, direct visual grounding and implicit visual referencing, and achieves consistent improvements in various LVLMs and hallucination benchmarks. 

\noindent \textbf{Limitations and future directions.} 
Although effective, FLB does not fundamentally resolve the structural limitations of training-free decoding. 
In particular, FLB cannot fully model context-dependent visual grounding as it injects visual evidence through the first-token logit rather than adapting to evolving visual semantics. 
Furthermore, it does not eliminate the underlying long-term decay introduced by RoPE, but instead mitigates its effect through early-token reinforcement. 
Extending FLB with adaptive token-level visual grounding or decay-aware positional mechanisms would be interesting future directions. 

\section*{Acknowledgements}
This work was supported by the DGIST Start-up Fund Program of the Ministry of Science and ICT(2026010221), InnoCORE program of the Ministry of Science and ICT(26-InnoCORE-01), Institute of Information \& Communications Technology Planning \& Evaluation(IITP) grant funded by the Korea government(MSIT) (No.RS-2025-02219277, Al Star Fellowship Support (DGIST)), and 2026 innovation base artificial intelligence data convergence project project with the funding of the 2026 government (Ministry of Science and ICT) (S2201-26-1002).

{
    \small
    \bibliographystyle{ieeenat_fullname}
    \bibliography{main}
}

\clearpage
\setcounter{page}{1}
\maketitlesupplementary

\setcounter{section}{0}
\renewcommand{\thesection}{\Alph{section}}

\begin{figure}[t]
  \centering
   \includegraphics[width=0.9\linewidth]{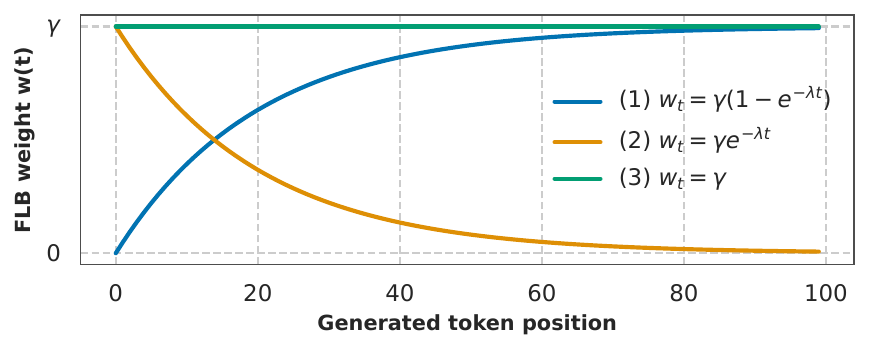}
    \caption{\textbf{FLB weight function candidates.} Comparison of three $w_t$ formulations for weight optimization across token positions.}
   \label{fig:calculation}
\end{figure}
\section{Detailed experiment result}
\label{supp:exp}
In all experiments, we used three hyperparameters, $\beta$, $\gamma$, and $\lambda$, which were set to 0.1, 0.3, and 0.05, respectively.
These values were chosen based on hyperparameter optimization experiments and were fixed throughout all evaluations.
In this supplementary section, we provide optimization results beginning with those presented in \cref{supp:hyp_opt}.
\section{Hyperparameter optimization}
\label{supp:hyp_opt}
\subsection{Weight function and hyperparameter optimization}
To find the best weight function in \cref{eq:FLB2}, 
we conducted an optimization experiment on different formulations of the weight equation using the AMBER generative task.
We evaluated three settings: exponentially (1) increasing, (2) decreasing, and (3) constant weight.
\cref{fig:calculation} illustrates the variation of the weight values across decoding steps in each setting.
We introduce a new metric to evaluate the performance of each weight defined as:
\begin{equation}
    \textit{object\_score} = 0.5((1-\textit{CHAIR})+\textit{Cover}),
    \label{eq:object_score}
\end{equation}
which combines the CHAIR and Cover scores, 
representing the proportion of hallucinated and ground truth object words 
in the generated sequences of the AMBER benchmark.
When evaluating performance according to this metric, $w_t = \gamma(1-e^{\lambda t})$ achieved the best performance as shown in \cref{tab:obj_scores_weight}.
This aligns with our intuition that as generated token position increases, the long-term decay problem becomes more severe, and hence a stronger suppression of long-term decay is required.
Consequently, we adopted this formula as the weight function of FLB. 

Furthermore, to identify the most effective values for $\gamma$ and $\lambda$, we conducted hyperparameter optimization.
We used the same evaluation metric as in the weight function optimization.
According to \cref{tab:obj_scores_gamma} and \cref{tab:obj_scores_instructblip}, both LLaVA1.5 and InstructBLIP models achieved the best performance when $\gamma = 0.3$ and $\lambda = 0.05$, which were used in the experiments.
Based on this result, we adopted these values as the hyperparameter settings for all subsequent experiments.
\begin{table}[t]
  \caption{Best object score in \cref{eq:object_score} at each weight function setting.}
  \label{tab:obj_scores_weight}
  \centering
  \small
    \begin{tabularx}{0.95\linewidth}{
    >{\centering\arraybackslash\hsize=0.24\linewidth}X|
    >{\centering\arraybackslash\hsize=0.18\linewidth}X
    >{\centering\arraybackslash\hsize=0.18\linewidth}X
    >{\centering\arraybackslash\hsize=0.2\linewidth}X
  }
    \toprule\toprule
    $w_t$& (1) inc. & (2) dec. & (3) const. \\
    \midrule
    object score & \textbf{72.3} & 72.2 &  71.75 \\
    \bottomrule\bottomrule
  \end{tabularx}
\end{table}
\begin{table}[t]
  \caption{Object score in \cref{eq:object_score} at each hyperparameter setting on LLaVA1.5.}
  \label{tab:obj_scores_gamma}
  \centering
  \small
    \begin{tabularx}{0.95\linewidth}{
    >{\centering\arraybackslash}X|
    >{\centering\arraybackslash}X
    >{\centering\arraybackslash}X
    >{\centering\arraybackslash}X
  }
    \toprule\toprule
    $\lambda$ / $\gamma$ & 0.1 & 0.3 & 0.5 \\
    \midrule
    0.005 & 71.15 & 71.2 & 71.4 \\
    0.02 & 71.65 & 71.8 & 72.1 \\
    0.05 & 71.8 & \textbf{72.3} & 71.7 \\
    \bottomrule\bottomrule
  \end{tabularx}
\end{table}
\begin{table}[t]
  \caption{Object score in \cref{eq:object_score} at each weight setting
  on InstructBLIP.}
  \label{tab:obj_scores_instructblip}
  \centering
  \small
    \begin{tabularx}{0.95\linewidth}{
    >{\centering\arraybackslash}X|
    >{\centering\arraybackslash}X
    >{\centering\arraybackslash}X
    >{\centering\arraybackslash}X
  }
    \toprule\toprule
    $\lambda$ / $\gamma$ & 0.1 & 0.3 & 0.5 \\
    \midrule
    0.005 & 72.4 & 72.3 & 72.0 \\
    0.02 & 72.3 & 72.4 & 72.3 \\
    0.05 & 72.2 & \textbf{72.6} & 72.1 \\
    \bottomrule\bottomrule
  \end{tabularx}
\end{table}
\subsection{Effect of \texorpdfstring{$\beta$}{beta} in adaptive plausible constraint}
To understand how the magnitude of adaptive plausible constraint affects the performance of FLB, we conducted a hyperparameter optimization over $\beta$.
\cref{tab:beta_opt} shows the results of the hyperparameter optimization for $\beta$.
A larger value of $\beta$ leads to more aggressive truncation, allowing only tokens with higher probabilities to be selected. 
We conducted this optimization on LLaVA-1.5 using the AMBER benchmark.
As shown in \cref{tab:beta_opt}, both excessively large and excessively small values of $\beta$ limit the performance improvement of FLB. 
This indicates that overly aggressive or insufficient truncation can negatively affect the model’s ability to suppress hallucinations while preserving correct information.

\begin{figure*}[!t]
\begin{mdframed}[
    backgroundcolor=gray!10,
    roundcorner=20pt,
]
\rule{\linewidth}{0.8pt}\vspace{5pt}
\textbf{Input:}\\
\includegraphics[width=0.4\textwidth]{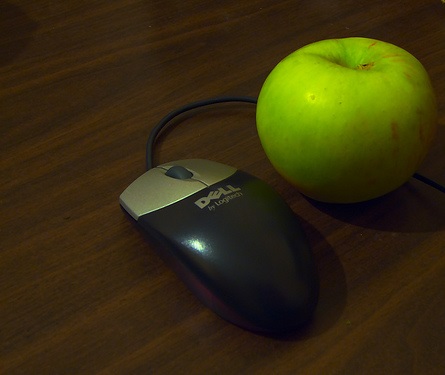}
\\
\text{[Prompt]} \\
Please describe this image in detail. \\
\\
\text{[First Logit Boosting with $\beta=0$]} \\
The scene features an Apple computer mouse and a green apple on a wooden table, with the mouse placed next to the apple on the right side. The apple is Apple's trademark in the image. The flat in the background indicates the mantle of the scene. The table's surface here is reflective and gray. The Sub In The Over The The The The The The The The The The The The The The The The The The The The The The The The The The The The The The The The The The The The The The The The The The The  ... \\
\\
\text{[First Logit Boosting with $\beta=0.1$]} \\
The image features a black and silver computer mouse placed next to a shiny green apple. The mouse is positioned on the left side of the table, while the apple is located on the right side, both sitting on the same wooden surface. This setting suggests a casual and comfortable workspace.
\\

\rule{\linewidth}{0.8pt}
\end{mdframed}
\captionof{figure}{Example outputs from LLaVA-1.5 with FLB applied using $\beta = 0$ and $\beta = 0.1$, illustrating that removing the adaptive plausible constraint ($\beta = 0$) leads to unnatural outputs.}
\label{fig:The_case}
\end{figure*}

In addition, when the adaptive plausible constraint is not applied (i.e., $\beta = 0$), 
we observed unnatural token selections, such as spurious insertions of tokens like ``The'' in the middle of a sentence, as illustrated in \cref{fig:The_case}. 
When the constraint is applied (e.g., $\beta = 0.1$), this erroneous behavior disappears, 
suggesting that the adaptive plausible constraint plays an important role in stabilizing token selection.
Based on these results, we selected 0.1 as the value of $\beta$, which achieves the highest object score and provides the best balance between reducing hallucinations and maintaining correct object coverage.
\subsection{Effect of decoding strategy}
In addition to the sampling-based decoding used in our main experiments, 
we also evaluated the performance under greedy decoding, 
where the token with the highest probability is selected at each step.
As shown in \cref{tab:greedy}, 
FLB consistently mitigates hallucination even when the decoding strategy is switched to greedy decoding, 
demonstrating that its effectiveness is robust to the choice of decoding method.

However, we also observed a slight decrease in Cover score under greedy decoding. 
This suggests that the deterministic nature of greedy decoding may restrict the FLB’s ability to fully describe ground truth objects in mentions, 
indicating that additional tuning or decoding adjustments may be necessary to further improve Cover score.

\begin{table*}
    \small
    \caption{Hyperparameter optimization on $\beta$ on AMBER benchmark.}
    \label{tab:beta_opt}
    \centering
    \begin{tabularx}{\textwidth}{
        >{\centering\arraybackslash}m{0.100\linewidth} |
        >{\centering\arraybackslash}m{0.15\linewidth}
        >{\centering\arraybackslash}m{0.15\linewidth}
        >{\centering\arraybackslash}m{0.15\linewidth}
        >{\centering\arraybackslash}m{0.15\linewidth} |
        >{\centering\arraybackslash}m{0.15\linewidth}
    }
    \toprule\toprule
    $\beta$ & \textbf{CHAIR\(\downarrow\)} &\textbf{Cover\(\uparrow\)} & \textbf{Hal\(\downarrow\)} & \textbf{Cog\ \(\downarrow\)} & \textbf{Object score\(\uparrow\)} \\ 
    \midrule
    0 & $7.8\,(\pm\,0.17)$ & $50.2\,(\pm\,0.41)$ & $39.7\,(\pm\,1.58)$ & $2.9\,(\pm\,0.08)$ & $71.2\,(\pm\,0.12)$\\
    0.01  & $7.5\,(\pm\,0.25)$  & $50.3\,(\pm\,0.5)$ & $37.5\,(\pm\,0.64)$ & $3.3\,(\pm\,0.17)$ & $71.4\,(\pm\,0.13)$\\ 
    0.05 & $6.8\,(\pm\,0.25)$ & $50.4\,(\pm\,0.19)$ & $33.3\,(\pm\,0.70)$ & $2.8\,(\pm\,0.17)$ & $71.8\,(\pm\,0.21)$\\
    0.1 & $6.1\,(\pm\,0.37)$ & $50.4\,(\pm\,0.22)$ & $31.4\,(\pm\,1.20)$ & $2.7\,(\pm\,0.24)$ & \textbf{72.1} ${(\pm\,0.12)}$ \\
    0.2 & $6.6\,(\pm\,0.21)$ & $50.2\,(\pm\,0.21)$ & $31.9\,(\pm\,1.07)$ & $2.9\,(\pm\,0.29)$ & $71.8\,(\pm\,0.18)$\\
    0.4 & $6.5\,(\pm\,0.08)$ & $50.5\,(\pm\,0.17)$ & $30.4\,(\pm\,0.52)$ & $3.3\,(\pm\,0.22)$ & $72.0\,(\pm\,0.10)$\\
    \bottomrule
    \bottomrule
    \end{tabularx}
\end{table*}

\begin{table*}[!t]
  \centering
  \small
  \caption{Performance comparison of LLaVA1.5 and greedy sampling on AMBER generative tasks. The \textbf{highest} scores are
marked in \textbf{bold}.}
\label{tab:greedy}
  \resizebox{\textwidth}{!}{%
    \begin{tabularx}{\textwidth}{
    >{\centering\arraybackslash}X|
    >{\centering\arraybackslash}X>{\centering\arraybackslash}X
    >{\centering\arraybackslash}X>{\centering\arraybackslash}X 
  }  
    \toprule\toprule

    \textbf{Method} & \textbf{CHAIR$\downarrow$} & \textbf{Cover$\uparrow$} & \textbf{Hal$\downarrow$} & \textbf{Cog$\downarrow$}\\
    \midrule
    Baseline & 7.1 & 50.5 & 32.4 & 3.8\\
    VCD     & 8.2 & 52.2 & 38.0 & 4.0\\
    ICD     & 6.4 & 51.0 & 30.6 & 3.2\\
    M3ID    & 7.0 & \textbf{55.8} & 37.5 & 2.8\\
    FLB (Ours) & \textbf{4.9} & 48.8 & \textbf{25.2} & \textbf{2.3} \\
    \bottomrule
    \bottomrule
  \end{tabularx}%
  }
    
\end{table*}
\begin{table*}[!t]
  \centering
  \small
  \caption{Comparison of the frequency of ``The'' appearing as the sentence-initial token with and without applying FLB.}
  \label{tab:The_ratio}
  \resizebox{\textwidth}{!}{%
    \begin{tabularx}{\textwidth}{
      >{\centering\arraybackslash}X|
      >{\centering\arraybackslash}X
      >{\centering\arraybackslash}X
      >{\centering\arraybackslash}X
      >{\centering\arraybackslash}X
      >{\centering\arraybackslash}X
      >{\centering\arraybackslash}X
    }
    \toprule\toprule
    \textbf{Method} & \textbf{Proportion of ``The'' among sentence-initial tokens} & \textbf{CHAIR$\downarrow$} & \textbf{Cover$\uparrow$} &  \textbf{Accuracy$\uparrow$} & \textbf{Detailedness$\uparrow$} & \textbf{Expression Diversity$\uparrow$} \\
    \midrule
    Baseline & 67.4\% & 11.9 & 49.6 & 4.83 & 4.65 & 5.36 \\
    FLB ($\gamma=0.1$) & 83.1\% & 7.5 & 51.1 & 6.21 & 5.80 & 6.23 \\
    FLB ($\gamma=0.3$) & 89.4\% & 5.7 & 50.3 & 6.47 & 5.91 & 6.34 \\
    FLB ($\gamma=0.5$) & 91.3\% & 5.8 & 49.2 & 6.26 & 5.77 & 6.10 \\
    FLB ($\gamma=0.7$) & 92.1\% & 5.3 & 48.7 & 6.29 & 5.85 & 6.22 \\
    \bottomrule
    \bottomrule
    \end{tabularx}%
  }
\end{table*}
\begin{table*}[t]
    \small
    \caption{Performance comparison of mPLUGOwl2 on AMBER. The \textbf{highest} scores are marked in \textbf{bold}.}
    \label{tab:mplugowl}
    \centering
    \begin{tabularx}{\textwidth}{
    >{\centering\arraybackslash}X|
    >{\centering\arraybackslash}X>{\centering\arraybackslash}X
    >{\centering\arraybackslash}X>{\centering\arraybackslash}X 
  }  
        \toprule\toprule
       \textbf{ Method} & \textbf{CHAIR\(\downarrow\)} & \textbf{Cover\(\uparrow\)} & \textbf{Hal\(\downarrow\)} & \textbf{Cog\(\downarrow\)} \\
        \midrule
        Baseline & 12.5 & 51.2 & 50.8 & 5.2 \\
        VCD & 11.3 & \textbf{53.1} & 46.4 & 5.5 \\
        FLB & \textbf{7.1} & 51.6 & \textbf{33.0} & \textbf{2.9} \\
        \bottomrule
        \bottomrule
    \end{tabularx}
    
\end{table*}
\begin{table*}[t]
  \caption{Performance comparison of LLaVA1.5 on discriminative tasks.}
  \label{tab:disc_result}
  \centering
  \small
  \resizebox{\textwidth}{!}{%
    \begin{tabularx}{\textwidth}{
    >{\centering\arraybackslash}X|
    >{\centering\arraybackslash}X>{\centering\arraybackslash}X
    >{\centering\arraybackslash}X>{\centering\arraybackslash}X
    >{\centering\arraybackslash}X>{\centering\arraybackslash}X|
    >{\centering\arraybackslash}X
  }  
    \toprule\toprule
    \multicolumn{1}{c}{\textbf{Benchmark}} &
    \multicolumn{6}{c}{\textbf{POPE}} &
    \multicolumn{1}{c}{\textbf{MME}} \\
    \cmidrule(lr){2-7} \cmidrule(lr){8-8}

    \multicolumn{1}{c}{\textbf{Dataset}} &
    \multicolumn{2}{c}{\textbf{Random}} &
    \multicolumn{2}{c}{\textbf{Popular}} &
    \multicolumn{2}{c}{\textbf{Adversarial}} \\

    \textbf{Metric} & \textbf{Acc.} & \textbf{F1} & \textbf{Acc.} & \textbf{F1}& \textbf{Acc.} & \textbf{F1} & \textbf{MME score} \\
    
    \midrule
    Baseline & 0.829 & 0.808 & 0.811 & 0.792
            & 0.786 & 0.771 &  114.20  \\
    $\beta$-only  & 0.846 & 0.826 &  0.827 & 0.809 
            & 0.801 & 0.786  & 115.88 \\
    FLB (Ours) & 0.846  &  0.826  &  0.827 & 0.809
            & 0.801 & 0.786 & 115.88 \\
    \bottomrule\bottomrule
  \end{tabularx}%
  }
\end{table*}
\section{Ratio of ``The'' in responses after applying FLB}
\label{sec:The_ratio}
Since FLB adds first logit, which assigns a high probability to ``The,'' to the logits of subsequent tokens,
we compare the proportion of ``The'' in the original responses and in the FLB applied responses using the 1,004 queries of the AMBER benchmark.
As shown in \cref{tab:The_ratio}, the proportion of ``The'' among all sentence-initial tokens rises by 21.9\%p
after applying FLB with $\gamma = 0.3$ and $\lambda = 0.05$, which is our optimized setting.
This raises concerns that FLB might reduce sentence diversity or lead to more monotonous phrasing. 
However, as demonstrated in \cref{subsec:QoS}, FLB does not degrade the overall quality of the generated sentences.

To further examine this effect, we conducted an additional GPT-4V based evaluation comparing baseline LLaVA-1.5 and FLB outputs across different $\gamma$ values.
Because $\gamma$ determines the maximum strength of the weighting function, we evaluated FLB with $\gamma \in \{0.1, 0.3, 0.5, 0.7\}$, where $\gamma = 0.3$ corresponds to our optimized setting.
The prompt format used for this evaluation is shown in \cref{fig:prompt_temp2}.
As shown in \cref{tab:The_ratio}, FLB consistently achieves high GPT-4V evaluation scores across all tested $\gamma$ values. 
Notably, FLB also outperforms the baseline in the newly introduced \textit{Expression Diversity} metric, demonstrating that the increase in “The” does not negatively impact stylistic variety or expression richness.
If an application requires reducing the proportion of “The,” this can be easily controlled by lowering the value of $\gamma$. 
Overall, these results indicate that FLB does not harm linguistic diversity or sentence quality despite influencing the distribution of sentence-initial tokens.
\section{{Other backbone models}}
\label{supp:otherbench}
{
In the main paper, we evaluate FLB using LLaVA1.5 and InstructBLIP as LVLM backbones. 
To further assess its generalization across different models, we additionally conduct experiments on mPLUG-Owl2 using the AMBER benchmark. 
All hyperparameters are kept identical to those used in the main experiments. 
As shown in~\cref{tab:mplugowl}, consistent with the results in the main paper, FLB outperforms both the baseline and VCD across hallucination-related metrics.
These results further demonstrate the robustness and generalizability of FLB across different LVLM backbones.}
\section{{Results on discriminative task}}
\label{supp:discriminative}
{
FLB is designed to mitigate hallucination that becomes more pronounced in longer generated sequences. 
Therefore, its impact is expected to be limited in discriminative tasks with short outputs. 
We evaluate FLB on two representative object hallucination benchmarks, POPE~\cite{li2023evaluating} and MME~\cite{fumme}, using LLaVA1.5 as the backbone. 
We compare FLB with baseline decoding and a $\beta$-only setting, which applies only the adaptive plausibility constraint without logit boosting.
Other experimental settings are identical to those in the main paper.
As shown in~\cref{tab:disc_result}, applying FLB yields the same performance as $\beta$-only decoding, indicating minimal effect in this setting.
This suggests that FLB is particularly beneficial for generative settings with longer outputs.
}
\section{Prompt of GPT-4V evaluation}
\label{supp:gpt4v}
To assess the quality of responses after applying FLB, 
we use GPT-4V to evaluate both accuracy and level of detail in the generated outputs.
The specific prompt format used for this evaluation is shown in \cref{fig:prompt_temp}, and an example of the resulting evaluation is provided in \cref{fig:GPT4_ex}.
For the analysis in \cref{sec:The_ratio}, we additionally include the corresponding prompt format and evaluation example in \cref{fig:prompt_temp2} and \cref{fig:GPT4_ex2}, respectively.

\clearpage
\begin{tcolorbox}[
    float*=t,                               
    width=\textwidth,
    colback=gray!10,                        
    colframe=black,                       
    rounded corners,                      
    boxrule=0.8pt,                       
    enlarge top by=5pt,
    enlarge bottom by=5pt,
    breakable                               
]
\rule{\linewidth}{0.8pt}\vspace{5pt}
\textbf{Description:} \\
You are required to score the performance of two AI assistants in describing a given image.\\
Pay special attention to hallucination — parts of descriptions inconsistent with the image content\\
(e.g., objects not present, wrong counts, colors, or positions).\\
\\
Rate each response on a scale of 1 to 10 for:\\
1. Accuracy — whether the response matches the image content.\\
2. Detailedness — whether the response is rich in necessary details (but not hallucinated).\\
\\
\textbf{Input format:} \\
\\
\text{[Assistant 1]} \\
\text{\{Response 1\}} \\
\text{[End of Assistant 1]} \\
\\
\text{[Assistant 2]} \\
\text{\{Response 2\}} \\
\text{[End of Assistant 2]} \\
\\
\textbf{Output format:} \\
\\
Accuracy: [score1] [score2] \\
Reason:\\
Detailedness: [score1] [score2]\\
Reason: \\
\rule{\linewidth}{0.8pt}
\captionof{figure}{\textbf{Prompt template for the GPT-4V evaluation}. Assistant 1 and Assistant 2 indicate the response of baseline method and FLB method, respectively. \{Response 1\} is the baseline response, and \{Response 2\} is the response after applying FLB.}
\label{fig:prompt_temp}
\end{tcolorbox}
\clearpage

\begin{tcolorbox}[
    float*=t,                              
    width=\textwidth,
    colback=gray!10,                        
    colframe=black,                         
    rounded corners,                        
    boxrule=0.8pt,                         
    enlarge top by=5pt,
    enlarge bottom by=5pt,
    breakable                               
]

\rule{\linewidth}{0.8pt}\vspace{5pt}

\textbf{Input:}\\[4pt]

\includegraphics[width=0.25\textwidth]{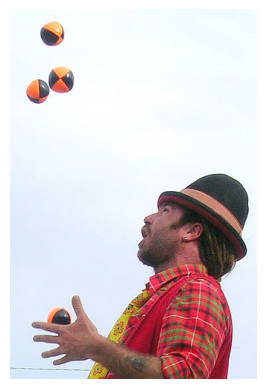}

\vspace{6pt}

\textbf{[Assistant 1]} \\
In the image, a man dressed in a top hat and scarf is balancing four orange balls on his fingers.\\
He appears to be performing a daring and skillful act of keeping the balls in place while holding only one finger.\\
The man seems to be confident in his presentation, capturing the attention of those watching.\\
The four balls are placed at different positions on his fingers, with each ball spaced apart from each other.\\
The arrangement challenges his dexterity and hand-eye coordination, highlighting his abilities as an entertainer.\\
\textbf{[End of Assistant 1]}\\[6pt]

\textbf{[Assistant 2]}\\
The image features a man wearing a top hat and a red and yellow plaid shirt, standing in the middle of the scene.\\
He is in the midst of juggling three orange and black balls in the air, showcasing his skill and coordination.\\
The man is the main focus of the image, and the scene captures the excitement and concentration of the juggling act.\\
\textbf{[End of Assistant 2]}\\[10pt]

\textbf{Output:}\\[4pt]

Accuracy: 1 9\\
- Assistant 1: Incorrectly describes the man as balancing the balls on his fingers, which is not the case. He is juggling four balls.\\
- Assistant 2: Accurately describes the man juggling, but mistakenly mentions three balls when there are actually four.\\[8pt]

Detailedness: 3 7\\
- Assistant 1: Provides incorrect information with unnecessary details about balancing, which reduces relevance.\\
- Assistant 2: Offers relevant details about the attire and the juggling act, despite the minor error in the number of balls.\\

\rule{\linewidth}{0.8pt}
\captionof{figure}{\textbf{Example of a GPT-4V evaluation}. Assistant 1 and Assistant 2 indicates the response of baseline methd and FLB method, respectively. The response generated with FLB obtains a higher evaluation score.}
\label{fig:GPT4_ex}
\end{tcolorbox}
\clearpage
\begin{tcolorbox}[
    float*=t,                               
    width=\textwidth,
    colback=gray!10,                        
    colframe=black,                         
    rounded corners,                        
    boxrule=0.8pt,                          
    enlarge top by=5pt,
    enlarge bottom by=5pt,
    breakable                               
]
\rule{\linewidth}{0.8pt}\vspace{5pt}
\textbf{Description:} \\
You are required to score the performance of two AI assistants in describing a given image.\\
Pay special attention to hallucination — parts of descriptions inconsistent with the image content\\
(e.g., objects not present, wrong counts, colors, or positions).\\
\\
Rate each response on a scale of 1 to 10 for:\\
1. Accuracy — whether the response matches the image content.\\
2. Detailedness — whether the response is rich in necessary details (but not hallucinated).\\
3. Expression Diversity — whether the wording is varied and natural, avoiding repetitive or formulaic phrasing while remaining clear.\\
\\
\textbf{Input format:} \\
\\
\text{[Assistant 1]} \\
\text{\{Response 1\}} \\
\text{[End of Assistant 1]} \\
\\
\text{[Assistant 2]} \\
\text{\{Response 2\}} \\
\text{[End of Assistant 2]} \\
\\
\text{[Assistant 3]} \\
\text{\{Response 3\}} \\
\text{[End of Assistant 3]} \\
\\
\text{[Assistant 4]} \\
\text{\{Response 4\}} \\
\text{[End of Assistant 4]} \\
\\
\text{[Assistant 5]} \\
\text{\{Response 5\}} \\
\text{[End of Assistant 5]} \\
\\
\textbf{Output format:} \\
\\
Accuracy: [score1] [score2] [score3] [score4] [score5] \\
Reason:\\
Detailedness: [score1] [score2] [score3] [score4] [score5] \\
Reason: \\
Expression Diversity: [score1] [score2] [score3] [score4] [score5] \\
Reason: \\
\rule{\linewidth}{0.8pt}
\captionof{figure}{\textbf{Prompt template for the additional GPT-4V evaluation}. Assistant 1 indicates the baseline response, while Assistant 2–5 correspond to FLB responses with $\gamma \in {0.1, 0.3, 0.5, 0.7}$, respectively.
\{Response 1\} denotes the baseline response, while \{Response 2\}, \{Response 3\}, \{Response 4\}, and \{Response 5\} correspond to the responses generated with FLB under different $\gamma$ settings.}
\label{fig:prompt_temp2}
\end{tcolorbox}
\clearpage

\begin{tcolorbox}[
    float*=t,                               
    width=\textwidth,
    colback=gray!10,          
    colframe=black,                
    rounded corners,                     
    boxrule=0.8pt,                        
    enlarge top by=5pt,
    enlarge bottom by=5pt,
    breakable                             
]

\rule{\linewidth}{0.8pt}\vspace{5pt}

\textbf{Input:}\\[4pt]

\includegraphics[width=0.3\textwidth]{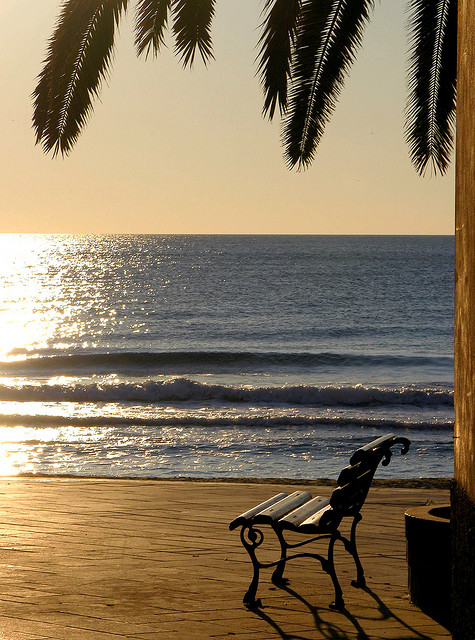}

\vspace{6pt}

\textbf{[Assistant 1]} \\
\textbf{The} image depicts a serene beach scene with benches and a sunset as the main focal point. \\
Two wooden benches face the ocean, one closer to the left side of the frame and the other towards the right. \\
\textbf{The} sun is setting, casting a warm glow on the scene. \\
In addition to the benches, there are palm trees visible in the background, making the location seem more relaxing and tropical. \\
They frame the view while a person can be seen standing a bit further back on the beach, gazing towards the ocean and enjoying the picturesque atmosphere.\\
\textbf{[End of Assistant 1]}\\[6pt]

\textbf{[Assistant 2]}\\
\textbf{The} image features a picturesque beach scene with the ocean in the background. \\
A bench is placed near the shoreline, offering a view of the water. \\
Two benches are visible, one positioned closer to the left side of the frame and the other closer to the right side.\\ They are arranged to face the water, creating a comfortable spot for beachgoers to relax and enjoy the view. \\
\textbf{The} benches appear to be empty in this scene.\\
\textbf{[End of Assistant 2]}\\[6pt]

\textbf{[Assistant 3]} \\
\textbf{The} image features a beach with a bench facing the ocean, creating a serene atmosphere. \\
\textbf{The} bench is positioned under the shade of a palm tree, providing a comfortable spot for visitors to relax and enjoy the beautiful view. \\
\textbf{The} ocean in the background is calm, with waves gently lapping on the shore. \\
\textbf{Th}e setting sun adds a touch of warmth and tranquility to the scene, making it a perfect location for a peaceful day by the sea.\\
\textbf{[End of Assistant 3]}\\[6pt]

\textbf{[Assistant 4]} \\
\textbf{The} image showcases a serene beach scene at sunset with a beautiful ocean view. \\
In the foreground, a bench is situated near the water, overlooking the ocean and the sunset. \\
\textbf{The} bench is located on a wooden walkway that extends along the beach, providing a picturesque view of the ocean and the sun setting in the background. \\
\textbf{The} bench is the focal point of the scene, inviting visitors to sit and enjoy the captivating beauty of the sunset and the tranquility of the beach.\\
\textbf{[End of Assistant 4]}\\[6pt]

\textbf{[Assistant 5]} \\
\textbf{The} image depicts a serene beach scene with a bench situated near the water's edge, overlooking the ocean. \\
\textbf{The} bench is placed on the beach and is located in the center of the image. \\
\textbf{The} beach is lined with a few trees, and the sun is shining on the scene, creating a peaceful and inviting atmosphere. \\
\textbf{The} bench provides a comfortable spot for visitors to sit and enjoy the ocean view.\\
\textbf{[End of Assistant 5]}\\[10pt]

\textbf{Output:}\\[4pt]

Accuracy: 2 7 9 5 4\\
- Assistant 1: Contains inaccuracies, like stating there are two benches and mentioning a person, which is not visible. \\
- Assistant 2: Correctly identifies one bench but incorrectly describes two benches when only one is visible.\\
- Assistant 3: Accurately describes a single bench and palm trees but doesn’t mention any incorrect elements.\\
- Assistant 4: Incorrectly describes a wooden walkway and implies proximity to water, which isn't accurate.\\
- Assistant 5: Correct in describing the bench and ocean, but inaccurately mentions trees lining the beach.\\[6pt]

Detailedness: 3 6 8 6 5\\
- Assistant 1: Includes imaginary elements (second bench, person), compromising detail accuracy.\\
- Assistant 2: Provides a moderate level of detail but errs in number of benches.\\
- Assistant 3: Offers rich and correct details about the scene and environment.\\
- Assistant 4: Attempts detailed storytelling but uses elements not present in the image (wooden walkway).\\
- Assistant 5: Lacks some specific details and includes inaccuracies (tree lineup).\\[6pt]

Expression Diversity: 4 6 8 7 6\\
- Assistant 1: Uses some varied language but hallucinates too much, reducing quality.\\
- Assistant 2: Contains some varied expressions but redundant in mistaken mention of two benches.\\
- Assistant 3: Well-varied expressions with descriptive phrases effectively painted a vivid scene.\\
- Assistant 4: Although somewhat expressive, incorrect details detracted from overall diversity.\\
- Assistant 5: Fair use of diverse expressions, lessened by inclusion of inaccurate content.\\[6pt]

\rule{\linewidth}{0.8pt}
\captionof{figure}{\textbf{Example of an additional GPT-4V evaluation}. Responses generated with FLB under different $\gamma$ values receive different evaluation scores. Assistant 1 shows the baseline response, while Assistant 2–5 correspond to FLB outputs with $\gamma \in {0.1, 0.3, 0.5, 0.7}$, respectively. \textbf{Bold} highlights occurrence of “The”. After applying FLB, the frequency of ``The'' increases, but the overall response quality does not decrease.}
\label{fig:GPT4_ex2}
\end{tcolorbox}

\end{document}